%% file: main.tex
\documentclass[journal]{IEEEtran}
\usepackage{amsmath,amsfonts}
\usepackage{algpseudocode}
\usepackage{algorithm}
\usepackage{array}
\usepackage[caption=false,font=footnotesize]{subfig}
\usepackage{textcomp}
\usepackage{stfloats}
\usepackage{url}
\usepackage{verbatim}
\usepackage{graphicx}
\usepackage{cite}
\usepackage{bibentry}
\usepackage{multirow}
\usepackage{amssymb}
\usepackage{tabularx}
\usepackage{xcolor}
\usepackage{wrapfig}
\usepackage{booktabs}
\usepackage{amsthm}
\usepackage{capt-of}

\input{math_commands.tex}

\begin{document}

\title{Heterogeneous-Modal Unsupervised Domain Adaptation via Latent Space Bridging}

\author{Jiawen~Yang,
        Shuhao~Chen,
        Shengtao~Zhang, Ke~Tang,~\IEEEmembership{Fellow,~IEEE}, 
        Yu~Zhang,~\IEEEmembership{Member,~IEEE}, 
        and Qiang Yang,~\IEEEmembership{Fellow,~IEEE}%
\thanks{J. Yang is with Department of Computer Science and Engineering, Southern University of Science and Technology, Shenzhen, China and Faculty of Computer Science and Artificial Intelligence, Shenzhen University of Advanced Technology, Shenzhen, China (e-mail: yangjwen55@gmail.com).}%
\thanks{S. Chen is with Department of Computer Science and Engineering, Southern University of Science and Technology, Shenzhen, China and Department of Computer Science and Engineering, The Hong Kong University of Science and Technology, Hong Kong, China (e-mail: schenht@connect.ust.hk).}%
\thanks{S. Zhang, K. Tang, and Y. Zhang are with the Department of Computer Science and Engineering, Southern University of Science and Technology, Shenzhen, China (e-mail: 12210730@mail.sustech.edu.cn, tangk3@sustech.edu.cn, yu.zhang.ust@gmail.com).}
\thanks{Q. Yang is with Department of Data Science and Artificial Intelligence, The Hong Kong Polytechnic University, Hong Kong, China (e-mail: profqiang.yang@polyu.edu.hk).}
\thanks{J. Yang and S. Chen contributed equally to this work.}
\thanks{Corresponding author: Yu Zhang.}
}

{}

\maketitle

\begin{abstract}
Unsupervised domain adaptation (UDA) effectively bridges the domain gap between a labeled source domain and an unlabeled target domain, but assumes that the two domains share the same modality. Heterogeneous domain adaptation (HDA) instead handles different feature spaces across domains, yet requires labeled target samples or paired data linking the source and target domains.
Neither applies when a labeled source domain and a fully unlabeled target domain each hold an entirely distinct modality (e.g., 2D images and 3D point clouds). 
To address this limitation, we introduce a new setting termed Heterogeneous-Modal Unsupervised Domain Adaptation (HMUDA), which transfers knowledge across modalities via an unlabeled bridge domain containing paired observations from both modalities, whose distribution may deviate from those of the source and target domains.  
To learn under the HMUDA setting, we propose Latent Space Bridging (LSB), a dual-branch framework where a feature consistency loss on paired bridge samples closes the modality gap and a class-centroid alignment loss reduces the source-target discrepancy. Extensive experiments on eight benchmark settings covering both 2D-to-3D and 3D-to-2D transfer demonstrate that LSB achieves state-of-the-art performance.
\end{abstract}


\section{Introduction}
\IEEEPARstart{U}{nsupervised} domain adaptation (UDA) methods \cite{yang2020transfer, gu2022unsupervised, zhao2020review} are effective at transferring knowledge from a labeled source domain to an unlabeled target domain, particularly when the two domains exhibit distributional differences.
While traditional UDA methods \cite{long2015learning, PL, luo2019taking, saltori2022cosmix, xiao2022transfer} have achieved notable success when the source and target domains share the same modality (e.g., the image modality), multi-modal UDA (MM-UDA, a.k.a.\ cross-modal UDA) \cite{xmuda_cvpr, DsCML, BFtD_XMUDA, zhang2022self, liang2025unidxmd, pan2026panda} further transfers knowledge between domains that both contain multiple modalities (e.g., images and 3D point clouds).

However, MM-UDA presumes that multi-modal data are available in both the source and target domains.
A significant challenge remains when each domain holds only a single and entirely distinct modality, such as transferring knowledge from labeled 2D images to unlabeled 3D point clouds. To see that, we take the semantic segmentation task \cite{guo2018review, strudel2021segmenter, li2022deep} in autonomous driving \cite{cheng20212, tang2020searching, yan20222dpass, zhuang2021perception, xu2025frnet} as an example.
Accurate segmentation of LiDAR 3D point clouds is critical for safe navigation, yet acquiring densely labeled 3D point clouds is expensive and time-consuming \cite{liu2021adversarial}. In contrast, large-scale 2D-image segmentation datasets~\cite{cordts2016cityscapes, zhou2017scene, caesar2018coco, neuhold2017mapillary} are abundant and relatively easy to annotate~\cite{kirillov2023segment}.
This asymmetry in annotation cost naturally motivates transferring knowledge from the label-rich 2D modality to the label-scarce 3D modality.
Meanwhile, vehicles equipped with synchronized cameras and LiDAR continuously record the same surrounding scenes from both sensors, so abundant \emph{unlabeled} paired 2D and 3D observations are produced as a by-product of the sensing system~\cite{xiao2024selective}.
The central question is therefore how to exploit this cheap paired resource to transfer knowledge across modalities without any 3D annotations.

One may consider heterogeneous domain adaptation (HDA)~\cite{day2017survey, fang2022semi, wang2022cross, yao2020discriminative, yao2021multisource, zhang2024rethinking} as a potential solution, since it also transfers knowledge across different feature spaces.
However, existing HDA methods are formulated as a two-domain problem in that they either require a few labeled target samples to anchor the adaptation~\cite{duan2012learning, wang2022cross} or exploit paired in-distribution data that is assumed to directly connect the source and target domains as auxiliary side information~\cite{dai2008translated, zhou2014hybrid}.
Neither assumption holds in our scenario where the target domain is fully unlabeled and the paired observations are not guaranteed to follow the distribution of either the source or the target domain.
Moreover, those methods learn from pre-extracted features \emph{without} an end-to-end learning manner for classification problems, and thus cannot produce the pointwise representations required by dense prediction tasks \cite{long2015fully, qi2017pointnet} such as semantic segmentation.

\begin{figure}[!t]
    \centering
    \includegraphics[width=1\linewidth]{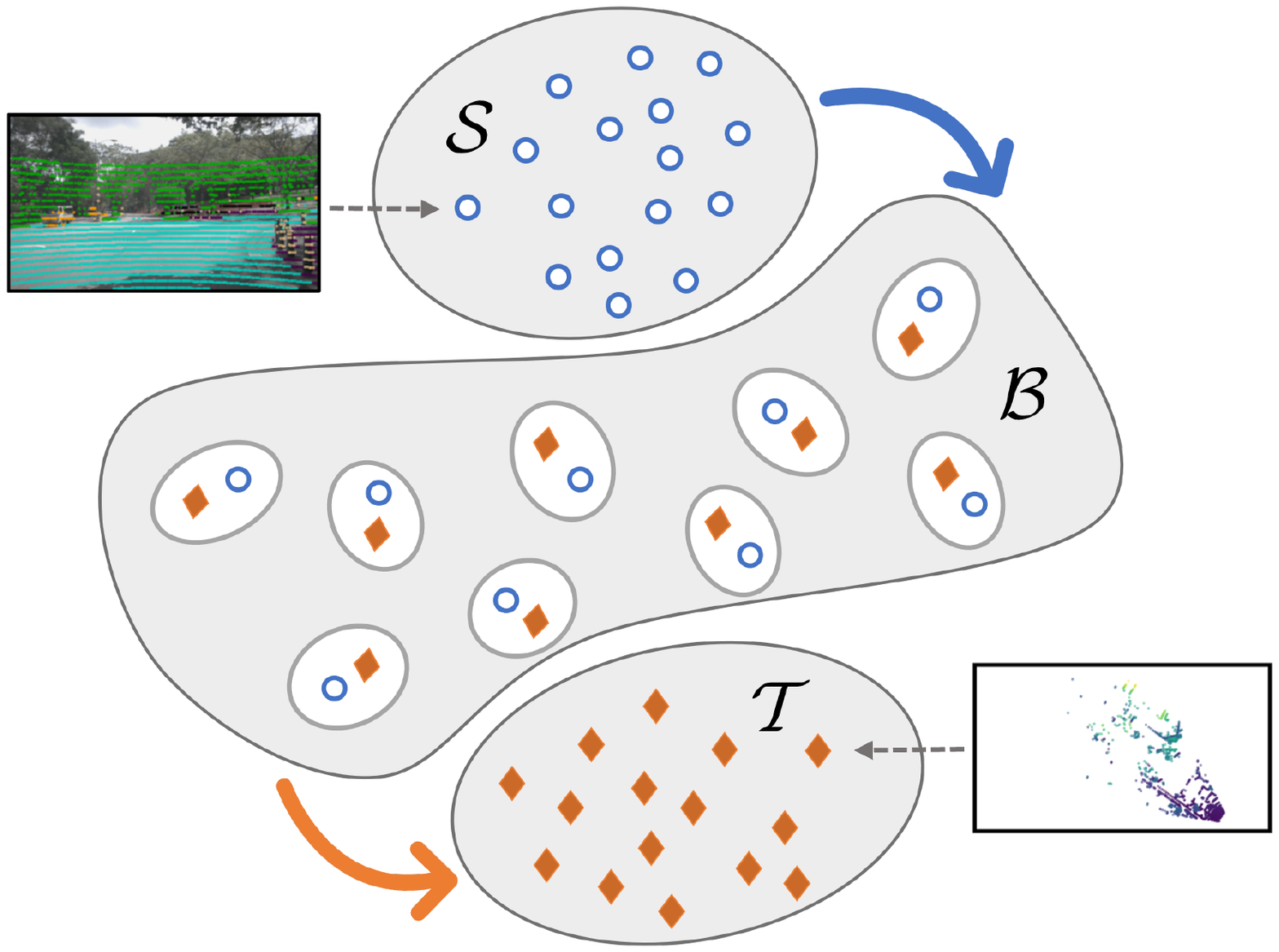}
    \caption{An illustration of the HMUDA setting. }
    \label{fig:Setting_image}
\end{figure}

These gaps motivate us to introduce a new setting, termed Heterogeneous-Modal Unsupervised Domain Adaptation (HMUDA).
Under HMUDA, labeled data are available only in a single-modality source domain (e.g., 2D images), while the target domain contains unlabeled data from an entirely different modality (e.g., 3D point clouds).
In addition, we assume access to an unlabeled \emph{bridge domain} that provides paired observations from both modalities, instantiating the cheap paired data resource described above.
Since labeled data in the source domain and paired data in the bridge domain are typically acquired through separate processes, the bridge domain is modeled as a third domain whose distribution may deviate from both the source and target, and the instance-level correspondence exists only within it.
Because paired samples in the bridge domain are collected under the same scene, their representations, despite residing in very different data spaces, should encode the same semantics, making the bridge domain a natural anchor for aligning the two modalities.
HMUDA therefore involves three distinct data domains: a labeled source domain, an unlabeled target domain, and an unlabeled bridge domain that provides cross-modal correspondence without being assumed to follow either the source or target distribution.
The HMUDA setting is illustrated in Figure~\ref{fig:Setting_image}, and Table~\ref{tbl:hmuda_comparsion} highlights its differences from existing domain adaptation (DA) paradigms, including UDA, MM-UDA, and HDA.

To learn in the HMUDA setting, we propose \underline{L}atent \underline{S}pace \underline{B}ridging (LSB), a novel framework specifically designed for semantic segmentation tasks.
The proposed LSB method employs a dual-branch architecture comprising a source network and a target network tailored for the source and target domains in different modalities, respectively.
The source network is trained with labeled source samples, while soft predictions produced by its EMA teacher on the source-modality bridge observations supervise the target network on their paired target-modality observations.
To enhance the feature alignment, we propose a \textbf{feature consistency loss} that encourages similar feature representations for paired samples from both modalities in the bridge domain, and a \textbf{domain alignment loss} that minimizes discrepancies between class centroids across the source and target domains.
Experimental results across various benchmark datasets demonstrate that the proposed LSB method effectively transfers knowledge from the source domain to the target domain, outperforming both the source-only baseline and existing UDA methods by a significant margin.

Our contributions are summarized as follows.
\begin{itemize}
\item We introduce heterogeneous-modal unsupervised domain adaptation, a new UDA setting that transfers knowledge across entirely different modalities by exploiting the instance-level correspondence within an unlabeled multi-modal bridge domain.
\item We propose Latent Space Bridging, a framework for semantic segmentation under HMUDA, to bridge the modality gap via a feature consistency loss on paired bridge samples and reduce the domain gap via a domain alignment loss between the source and target domains.
\item Extensive experiments on benchmark datasets demonstrate the effectiveness of the proposed LSB method, showcasing its ability to outperform existing methods.
\end{itemize}

\begin{table}[!t]
        \centering
        \caption{Comparison between HMUDA and existing DA settings. $\hM_1$ and $\hM_2$ denote two modalities. `Bridge' indicates whether a third unlabeled domain with paired modalities is available, and `E-to-E' indicates whether the specific model is trained end-to-end from raw inputs.}
        \resizebox{\linewidth}{!}{
        \begin{tabular}{l cc cc c c}
        \toprule
        & \multicolumn{2}{c}{Source} & \multicolumn{2}{c}{Target} & \multirow{2}{*}{Bridge} & \multirow{2}{*}{E-to-E} \\
        \cmidrule(lr){2-3} \cmidrule(lr){4-5}
        & $\hM_1$ & $\hM_2$ & $\hM_1$ & $\hM_2$ & & \\
        \midrule
        UDA & \cmark & \xmark & \cmark & \xmark & \xmark & \cmark \\
        MM-UDA & \cmark & \cmark & \cmark & \cmark & \xmark & \cmark \\
        HDA & \cmark & \xmark & \xmark & \cmark & \xmark & \xmark \\
        HMUDA & \cmark & \xmark & \xmark & \cmark & \cmark & \cmark \\
        \bottomrule
        \end{tabular}
        }
        \label{tbl:hmuda_comparsion}
\end{table}

\section{Related Work}

\subsection{Unsupervised Domain Adaptation}
UDA \cite{zhuang2020comprehensive} aims to transfer knowledge from a labeled source domain to an unlabeled target domain, addressing the challenge of the domain shift. 
Traditional UDA methods \cite{long2015learning, sun2016deep, tzeng2014deep, ganin2016domain} have achieved significant success in various applications by aligning the feature distributions between the source and target domains. 
For instance, discrepancy-based methods such as DAN \cite{long2015learning}, CORAL \cite{sun2016deep} and PPOT \cite{wang2024probability} minimize the statistical distance between the source and target feature distributions.
Meanwhile, adversarial-based methods like DANN \cite{ganin2016domain}, ADDA \cite{tzeng2017adversarial}, and CAPAN \cite{yu2024capan} employ adversarial training to learn domain-invariant features, leveraging a domain classifier to ensure that extracted features cannot be distinguished as originating from either the source domain or the target domain.

For the semantic segmentation task, UDA methods have been extended to tackle the pixel-level alignment challenge. Many of these methods also employ adversarial training for aligning the two domains. For instance, CyCADA \cite{hoffman2018cycada} uses cycle-consistent adversarial networks to adapt at both the pixel and feature levels. AdaptSegNet \cite{tsai2018learning} incorporates adversarial training at the output space to align the segmentation maps. CLAN \cite{luo2019taking} and CoSMix \cite{saltori2022cosmix} refine the adversarial approach by focusing on the class-level alignment and contextual consistency. 
DynAlign \cite{sun2025dynalign} further integrates UDA with foundation models to additionally bridge the label-level taxonomy gap across domains.
However, adversarial training can be unstable and incur a high computational burden for segmentation tasks \cite{mo2022review}. Different from them, the proposed LSB method uses discrepancy-based feature alignment, which directly minimizes the difference between source and target feature distributions.

\subsection{Multi-Modal Unsupervised Domain Adaptation}
MM-UDA \cite{xmuda_cvpr, DsCML, zhang2022self, wu2024clip2uda} transfers knowledge across domains where both the source and target domains contain multiple modalities, such as images and text or images and 3D point clouds. This approach leverages the complementary information from different modalities to enhance the adaptation performance. For example, 
Jaritz et al. introduce xMUDA \cite{xmuda_cvpr}, a cross-modal learning method that combines RGB images and LiDAR point clouds for improving the performance for 3D semantic segmentation.
DsCML \cite{DsCML} matches sparse 3D point features with dense 2D pixel features and adversarially aligns the cross-modal outputs across domains. 
SSE-xMUDA \cite{zhang2022self} presents a self-supervised exclusive learning mechanism that exploits the unique information of different modalities to complement each other. Dual-Cross \cite{li2022cross} designs a multi-modal stylized transfer module to alleviate the domain shift problem. 
UniDxMD \cite{liang2025unidxmd} discretizes 2D and 3D features via codebook-based quantization to obtain modality-agnostic representations, and PanDA \cite{pan2026panda} extends MM-UDA to multi-modal 3D panoptic segmentation via modality-degradation augmentation and dual-expert pseudo-label refinement.
These existing methods assume that both the source and target domains contain multi-modal data, and typically conduct the adaptation between the same type of multi-modal data. 
In contrast, our work introduces a novel setting, where both the source and target domains contain only one modality and the modalities in both domains are different. 

\subsection{Heterogeneous Domain Adaptation}
\label{sec:related_hda}
HDA addresses knowledge transfer between domains with different feature spaces or data types~\cite{day2017survey}.
Some HDA methods anchor the cross-space alignment on a few labeled target samples. For example, HFA~\cite{duan2012learning} learns a common space by maximizing cross-domain SVM margins, JGSA~\cite{zhang2017joint} optimizes coupled projections that reduce domain divergence while preserving discriminative structure, and CDSPP~\cite{wang2022cross} learns domain-specific projections preserving class structure.
CDFL \cite{lu2025heterogeneous} maximizes class-specific feature correlations across heterogeneous domains with selective pseudo-labeling, and Liu et al.~\cite{liu2024robust} project both domains into a union of multiple subspaces via a self-expressive model.
These methods require the existence of labeled target samples, 
and thus become infeasible when target labels are unavailable as in the proposed HMUDA setting.
Other HDA methods anchor on paired data, such as translated learning~\cite{dai2008translated} that estimates a feature-level translator from co-occurrence data and HHTL~\cite{zhou2014hybrid} that learns feature mappings from parallel instances between the source and target domains. 
However, they assume the paired data faithfully follow the source and target distributions, which rarely holds in practice. 
Moreover, both lines of methods focus on learning mappings over pre-extracted features and are tailored to classification tasks, and thus cannot produce the pointwise outputs required by dense prediction tasks in an end-to-end learning manner.
In contrast, the proposed HMUDA setting requires no target labels and treats the paired data as a separate bridge domain whose distribution may deviate from those of the source and target domains. Moreover, LSB trains modality-specific backbones end-to-end from raw inputs and directly produces pointwise predictions.

\section{Problem Formulation for HMUDA}
\label{sec:problem}
In this section, we introduce the HMUDA setting. Under the HMUDA setting, there exists a labeled source domain $\hS = \{ (\vx^s, \vy^s) \}$, where each sample consists of an input $\vx^s\in\mathcal{M}_1$ and pointwise segmentation labels $\vy^s \in \mathbb{R}^{C\times N}$ for $\vx^s$.
Here, $\mathcal{M}_1$ denotes the source modality of the input (e.g., the 2D image), $N$ is the number of labeled points in the input for semantic segmentation, and $C$ is the number of classes in the semantic segmentation task.
Moreover, there exists an unlabeled target domain $\hT=\{\vx^t\}$, where $\vx^t$ belongs to a different modality $\mathcal{M}_2$ (e.g., the 3D point cloud).
While our empirical study focuses on the modalities of 2D image and 3D point cloud, the HMUDA setting is modality-agnostic, where $\mathcal{M}_1$ and $\mathcal{M}_2$ can be instantiated by other heterogeneous modalities such as RGB image and infrared images, audio and text, or multi-spectral image and RGB image as long as the output space is consistent across domains.
To relate the two heterogeneous modalities in both domains, we assume the existence of an unlabeled bridge domain $\hB = \{ (\vx^{bs}, \vx^{bt}) \}$, where $\vx^{bs}\in \mathcal{M}_1$ and $\vx^{bt}\in \mathcal{M}_2$ are paired observations of the same scene from the two modalities.
Such paired data naturally exist in multi-sensor systems~\cite{xiao2024selective} (e.g., synchronized camera--LiDAR platforms). Because each pair $(\vx^{bs}, \vx^{bt})$ observes the same scene, the bridge domain serves as a natural anchor for aligning the two heterogeneous modalities.
Moreover, since the paired data in the bridge domain and the data in the source/target domain are acquired through separate processes, the bridge domain may differ from the source and target domains in terms of the data distribution.
In the HMUDA setting, we aim to transfer the knowledge in $\hS$ to help the learning of $\hT$ with the help of the bridge domain $\hB$.
 In summary, we give the definition for HMUDA.
\begin{definition} (\textbf{HMUDA})
In the heterogeneous-modal unsupervised domain adaptation setting, there are a labeled source domain \(\hS\) and an unlabeled target domain \(\hT\) across different modalities as well as an unlabeled bridge domain \(\hB\) that provides paired samples from both source and target modalities that have instance-level correspondence.
The objective of HMUDA is to leverage the labeled data in \(\hS\) and the cross-modal correspondence in \(\hB\) to improve the performance on the target domain \(\hT\).
\end{definition}

As shown in Table~\ref{tbl:hmuda_comparsion}, HMUDA differs from existing DA settings in terms of several key aspects.
UDA addresses domain shift between the source and target domains within the same modality, while MM-UDA extends this to multi-modal source and target domains but still assumes shared modalities across domains.
HDA handles heterogeneous feature spaces across domains, but existing HDA methods rely on pre-extracted features, anchored by a few labeled target samples or by paired data assumed to represent the source and target distributions.
By contrast, HMUDA addresses a more challenging scenario of transferring knowledge across entirely different modalities by end-to-end exploiting the instance-level correspondence available in the bridge domain without requiring any target labels.

In this paper, we study the HMUDA problem on the semantic segmentation task for three reasons. First, the annotation-cost asymmetry that motivates the HMUDA setting is pronounced in dense prediction tasks, where pointwise labels for 3D  point clouds are substantially more expensive to acquire than their 2D counterparts, and hence transferring knowledge without any target labels brings a big practical benefit. Second, the bridge domain arises naturally in this task, as synchronized and calibrated camera and LiDAR sensors provide paired observations whose correspondence holds at the point level. Third, semantic segmentation requires pointwise representations learned end-to-end from raw inputs, which is precisely the capability missing in existing HDA methods, making it a challenging testbed for HMUDA. 
Nevertheless, the HMUDA setting is not tied to the semantic segmentation task and can be naturally extended to other tasks such as skeleton-based action recognition from labeled RGB videos and emotion recognition from labeled text to unlabeled speech.

\begin{figure*}[t]
    \centering
    \includegraphics[width=1\linewidth]{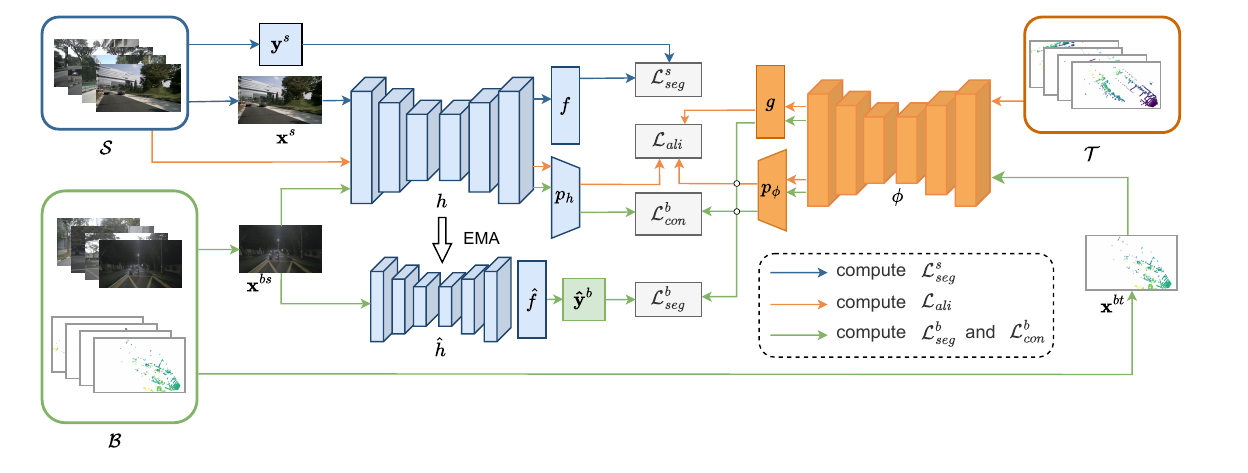}
    \caption{An illustration of the proposed LSB framework. 
    Lines within different colors denote the data flow for computing different losses.}
    \label{fig:architecture}
\end{figure*}

\section{Methodology}

In this section, we present the proposed LSB method to learn under the HMUDA setting. 
We first introduce the entire architecture of the proposed LSB method. 
Then, we introduce the losses in LSB to bridge the source and target domains by the bridge domain and align class centroids across domains. 
After giving the entire objective function and the algorithm for the LSB method, we analyze the error bound of the LSB method.

\subsection{Architecture}
\label{sec:architecture}
As depicted in Figure \ref{fig:architecture}, the proposed LSB method employs a dual-branch architecture to predict pointwise segmentation labels. 
The architecture in the LSB method consists of two distinct segmentation networks:  
(i) a \textbf{source network} $\{h, f\}$ tailored for the source modality $\hM_1$ with an encoder $h(\cdot): \hM_1 \rightarrow \mathbb{R}^{d_h \times N}$ and a classifier $f(\cdot): \mathbb{R}^{d_h \times N} \rightarrow \mathbb{R}^{C \times N}$, where $d_h$ is the output dimension of $h$ and $N$ is the number of points to be segmented as defined in Section~\ref{sec:problem};
(ii) a \textbf{target network} $\{\phi, g\}$ designed for the target modality $\hM_2$ with a feature extractor $\phi(\cdot):\hM_2 \rightarrow \mathbb{R}^{d_\phi \times N}$ and a classifier $g(\cdot): \mathbb{R}^{d_\phi \times N} \rightarrow \mathbb{R}^{C \times N}$, where $d_\phi$ is the output dimension of $\phi$ and $C$ is the number of classes in the segmentation task.

During training, the data in the source and bridge domains associated with modality $\hM_1$ are input into the source network, while the data in the target and bridge domains belonging to modality $\hM_2$ are input into the target network.
Hence, both the source and target networks can predict the pointwise segmentation labels independently with inputs of different modalities. 

\subsection{Bridging Source and Target Networks}
\label{sec:bridging_model}

To train the source network, a natural solution is to minimize the cross-entropy loss between the prediction of input $\vx^s$ (i.e., $f(h(\vx^s))$) and its corresponding labels $\vy^s$. Recall from Section~\ref{sec:problem} that $\vy^s$ is a $C \times N$ matrix, whose $i$-th column $\vy^s_{:,i}$ is the one-hot label vector over the $C$ classes for the $i$-th point of $\vx^s$. This segmentation loss, denoted by $\mathcal{L}_{\text{seg}}^{s}$, is formally defined as
\begin{equation}
\label{eqn:seg_loss_s}
\mathcal{L}_{\text{seg}}^{s} (\vx^s, \vy^s) = -\frac{1}{N} \sum_{i=1}^{N} (\vy^s_{:,i})^\top \log\big(f(h(\vx^s))_{:,i}\big),
\end{equation}
where $f(h(\vx^s))_{:,i}$ denotes the $i$-th column of the prediction, i.e., the predicted class distribution of the $i$-th point.

Since the label information for the target domain is unavailable, directly training the target network in the way of the source domain is infeasible.
To address this issue, we use the bridge domain instead to train the target network. Specifically, 
we introduce a teacher source model, comprising $\hat{f}$ and $\hat{h}$, which are updated from $f$ and $h$ using the exponential moving average (EMA)~\cite{tarvainen2017mean} at each iteration as
\begin{align}
\label{eqn:ema}
    \vtheta_{\hat{f}} \gets \alpha \vtheta_{\hat{f}} + (1-\alpha) \vtheta_{f}, \  \vtheta_{\hat{h}} \gets \alpha \vtheta_{\hat{h}} + (1-\alpha) \vtheta_{h},
\end{align}
where $\vtheta_f$, $\vtheta_h$, $\vtheta_{\hat{f}}$ and $\vtheta_{\hat{h}}$ denotes the parameters of $f$, $h$, $\hat{f}$, $\hat{h}$, respectively. $\alpha$ is a hyperparameter that is dynamically adjusted during training as $\alpha = \min\left(\frac{t}{t + 1}, \alpha\right)$, where $t$ denotes the number of training iterations completed so far. 
For a sample pair $(\vx^{bs}, \vx^{bt})$ in the bridge domain $\hB$, the teacher source network produces the soft class-probability distribution
\begin{equation}
    \hat{\vq}^b = \operatorname{softmax}\!\left(\hat{f}(\hat{h}(\vx^{bs}))\right).
\end{equation}
We use this distribution as a soft supervisory signal for the paired target-modality observation. Specifically, the bridge segmentation loss is
\begin{align}
\label{eqn:seg_loss_b}
\mathcal{L}_{\text{seg}}^{b} (\vx^{bs}, \vx^{bt})
&= \frac{1}{N} \sum_{i=1}^{N}
D_{\mathrm{KL}}\!\left(
\hat{\vq}^b_{:,i}\,\middle\|\,
\operatorname{softmax}\!\left(g(\phi(\vx^{bt}))_{:,i}\right)
\right),
\end{align}
where $\hat{\vq}^b_{:,i}$ is the EMA teacher's class-probability distribution for the $i$-th point. This soft-target formulation preserves the teacher's uncertainty across classes and matches the implementation based on KL divergence.

For an input sample pair $(\vx^{bs}, \vx^{bt})$, the source and target networks should extract similar features since they share the same labels.
To encourage consistent features between the source and target networks, we introduce projection networks $p_h: \mathbb{R}^{d_h \times N} \rightarrow \mathbb{R}^{d \times N}$ and $p_\phi: \mathbb{R}^{d_\phi \times N} \rightarrow \mathbb{R}^{d \times N}$, which map the source and target features into a $d$-dimensional shared feature space, respectively.
To minimize the discrepancy between the projected features of the source and target networks, we define the feature consistency loss $\hL_{\text{con}}^b$ as
\begin{align}
\hL_{\text{con}}^b(\vx^{bs}, \vx^{bt}) = \frac{1}{Nd} \sum_{i=1}^N
&\left\|p_h(h(\vx^{bs}))_{:,i}-p_\phi(\phi(\vx^{bt}))_{:,i}\right\|_2^2 \nonumber\\
&+\lambda_w \|\vw\|_2^2, \label{eqn:consistent_loss}
\end{align}
where $\|\cdot\|_2$ denotes the $\ell_2$ norm of a vector and $\lambda_w>0$ is a hyper-parameter controlling the strength of the regularization term $||\vw||_2^2$ with $\vw$ including parameters in $p_h$ and $p_\phi$.

\subsection{Cross-Modal Domain Alignment}
\label{sec:domain_alignment}
Directly training the models using the source and bridge domain often leads to overfitting\cite{tzeng2014deep}, resulting in diminished performance on the target domain. 
To alleviate this problem, inspired by the discrepancy-based UDA methods~\cite{long2015learning, zhu2020deep}, we learn the representation that minimizes the distance between the source and target domains.
To begin with, we obtain the pseudo label $\hat{\vy}^t$ for each $\vx^t \in \hT$ by the target network. 
For class $c$, we define the class centroids $\vm^s_c \in \mathbb{R}^{d}$ of the source domain and $\vm^t_c \in \mathbb{R}^{d}$ of the target domain as
\begin{align}
    \vm^s_c &= \left( \sum_{ (\vx^s, \vy^s) \in {\hS} }\vy_c^s \mathbf{1} \right)^{-1}  \sum_{ (\vx^s, \vy^s) \in {\hS} } \!\! p_h(h(\vx^s)) (\vy_c^s)^\top\label{centroid_source_domain}\\
    \vm^t_c &= \left({ \sum_{ \vx^t \in {\hT} } \hat{\vy}_c^t \mathbf{1}}\right)^{-1} {\sum_{ \vx^t \in {\hT} } p_{\phi}  (\phi(\vx^t)) (\hat{\vy}_c^t)^\top},\label{centroid_target_domain}
\end{align}
where $\vy_c^s$ and $\hat {\vy}_c^t$ denote the $c$-th row of $\vy^s$ and $\hat{\vy}^t$, respectively, and $\mathbf{1}$ denotes a $N$-dimension vector where all elements equal to one.
We expect the source and target models to share similar centroids for each class after the projection networks.
To this end, we minimize the discrepancy between the centroids using the alignment loss $\hL_{\text{ali}}$ as
\begin{equation}
\label{eqn:alignment_loss}
    \hL_{\text{ali}} ({\hS}, {\hT}) = \frac{1}{C} \sum_{c=1}^{C} 1 - \text{cos}\big( \vm^s_c, \vm^t_c \big),
\end{equation}
where $\text{cos}(\cdot)$ denotes the cosine similarity function. 
Since each centroid averages the features of all points assigned to a class, 
errors in individual pseudo labels are largely averaged out in $\vm^t_c$. 
By pulling the class centroids of the two domains together, $\hL_{\text{ali}}$ 
directly reduces the source-target discrepancy and prevents the target network 
from overfitting to the bridge distribution.

\subsection{Objective Function and Algorithm}
\label{sec:algorithm}

We jointly learn the source and target networks as well as the projection networks by minimizing the final objective $\hL(\hS, \hB, \hT)$, which combines $\hL_{\text{seg}}^s$, $\hL_{\text{seg}}^b$, $\hL_{\text{con}}$ and $\hL_{\text{ali}}$ together as 
\begin{align}
    \hL(\hS, \hB, \hT) = \!\!\!\! & \sum_{(\vx^s, \vy^s) \in \hS} \Big( \hL_{\text{seg}}^s (\vx^s, \vy^s) \Big) +  \lambda_{\text{a}} \hL_{\text{ali}}({\hS}, {\hT}) \label{eqn:total_loss}  \\
     & + \!\!\!\!\!\! \sum_{(\vx^{bs}, \vx^{bt}) \in \hB} \!\!\! \Big( \hL_{\text{seg}}^b(\vx^{bs}, \vx^{bt}) + \lambda_{\text{c}} \hL_{\text{con}}^b(\vx^{bs}, \vx^{bt}) \Big),\nonumber
\end{align}
where $\lambda_{\text{c}}>0$ and $\lambda_{\text{a}}>0$ are hyperparameters to balance different losses. 
The overall algorithm for the LSB method is provided in Algorithm \ref{app:overall_algorithm}.

\begin{algorithm}
\caption{The LSB method.}
\label{app:overall_algorithm}
\begin{algorithmic}[1]
\Require{Source domain ${\hS}$, target domain ${\hT}$, bridge domain $\hB$, learning rate $\eta$, hyperparameters $\lambda_c$,  $\lambda_a$, and $\lambda_w$; trainable parameters $\vtheta$: source network $\{h, f\}$, target network $\{\phi, g\}$ and projection networks $\{p_h, p_\phi\}$;}
\State Initialize the source teacher model $\{\hat{h},\hat{f}\}$.
\Repeat
\State Sample a batch of data $\widetilde{\hS} \subset \hS, \widetilde{\hB} \subset \hB, \widetilde{\hT} \subset \hT$.
\For{$(\vx^s, \vy^s) \in \widetilde{\hS}$}
    \State Compute loss $\hL_{\text{seg}}^s (\vx^s, \vy^s)$ by Eq.~(\ref{eqn:seg_loss_s}).
    \EndFor
\For{$(\vx^{bs}, \vx^{bt}) \in \widetilde{\hB}$}
    \State Compute loss $\hL_{\text{seg}}^b(\vx^{bs}, \vx^{bt})$ by Eq.~(\ref{eqn:seg_loss_b}).
    \State Compute loss $\hL_{\text{con}}^b(\vx^{bs}, \vx^{bt})$ by Eq.~(\ref{eqn:consistent_loss}).
\EndFor
\State Compute loss $\hL_{\text{ali}}(\widetilde{\hS}, \widetilde{\hT})$ by Eq.~(\ref{eqn:alignment_loss}).
\State Compute the overall loss $\hL$ by Eq.~(\ref{eqn:total_loss}).
\State $\vtheta \gets \vtheta - \eta \nabla_\vtheta \hL$;
\State Update $\{\hat{h},\hat{f}\}$ using EMA from $\{h, f\}$ by Eq.~\eqref{eqn:ema}.
\Until{convergence}
\end{algorithmic}
\end{algorithm}

\subsection{Analysis}
\label{sec:theoretical analysi}
In this section, we analyze the error bound under the HMUDA setting.
Let $\hE_s(\{h, f\})=\bE_{(\vx,\vy) \sim \hD_s}[f(h(\vx)) \neq \vy]$ and $\hE_t(\{\phi, g\}) = \bE_{(\vx, \vy) \sim \hD_t}[g(\phi(\vx)) \neq \vy]$ denote the expected error in the source domain with data distribution $\hD_s$ and target domain with data distribution $\hD_t$, respectively.
With the distribution $\hD_b$ of the bridge domain, let $\hE_{bs}(\{h, f\}) = \bE_{(\vx^{bs}, \vx^{bt}, \vy) \sim \hD_{b}}[f(h(\vx^{bs})) \neq \vy]$ and $\hE_{bt}(\{\phi, g\}) = \bE_{(\vx^{bs}, \vx^{bt}, \vy) \sim \hD_{b}}[g(\phi(\vx^{bt})) \neq \vy]$ be the expected errors corresponding to the source and target modalities within the bridge domain.
We denote the hypothesis space for the source network $\{h, f\}$ and the target network $\{\phi,g\}$ by $\hH_s$ and $\hH_t$, respectively. The optimized source and target networks are defined as follows.
\begin{definition}
    The ideal joint hypothesis for source and target modality is the hypothesis that minimizes the combined errors:
    \begin{align}
        \{h^\ast, f^\ast\} &= \argmin_{ \{h,f\} \in \hH_s} \hE_s(\{h,f\}) + \hE_{bs}(\{h,f\}) \\
        \{\phi^\ast, g^\ast\} &= \argmin_{ \{\phi,g\} \in \hH_t} \hE_{bt}(\{\phi, g\}) + \hE_{t}(\{\phi, g\}).
    \end{align}
\end{definition}

The combined error for source and target modality under the ideal hypothesis are denoted by
\begin{align}
    \lambda_s &= \hE_s(\{h^\ast,f^\ast\}) + \hE_{bs}(\{h^\ast,f^\ast\}) \nonumber \\ 
    \lambda_t & = \hE_{bt}(\{\phi^\ast, g^\ast\}) + \hE_{t}(\{\phi^\ast, g^\ast\}).
\end{align}
Moreover, similar to \cite{zhuang2024gradual}, we assume the error gap between modalities of the same domain is bounded by their feature gap as $\big| \hE_{bt} (\{\phi,g\}) - \hE_{bs}(\{h, f\}) \big| \leq L \bE_{(\vx^{bs}, \vx^{bt}) \sim \hD_b} \left[d(h(\vx^{bs}), \phi(\vx^{bt})) \right]$, where $|\cdot|$ denotes the absolute value, $L$ is a constant, and $d$ is a distance function (e.g., the $\ell_1$ distance). 
Then in the following theorem, we give a bound on the expected error of the target domain.

\begin{theorem}
\label{the:unpperboundProof}
For every $\{\phi, g\} \in \hH_t$, the expected error of the target domain is bounded as
\begin{align}
\label{eqn:target_bound}
\hE_{t} (\{\phi,g\})  \leq & \hE_{s} (\{h, f\})  + L \bE_{(\vx^{bs}, \vx^{bt}) \sim \hD_b} \left[d(h(\vx^{bs}), \phi(\vx^{bt})) \right] \nonumber \\
& + \frac{1}{2} d_{\hH_s \Delta \hH_s} (\hD_s, \hD_{b})  + \frac{1}{2}  d_{\hH_t \Delta \hH_t } (\hD_{b}, \hD_{t}) \nonumber \\
&  + (\lambda_s + \lambda_t),
\end{align}
where $d_{\hH_s \Delta \hH_s}$ and $d_{\hH_t \Delta \hH_t }$ are the $\hH \Delta \hH$-distances~\cite{ben2010theory} between domains in the source and target modalities, respectively.
\end{theorem}

\label{app:proof_upbound}
\begin{proof}
Based on Theorem 2 of \cite{ben2010theory} applied to the source and bridge domains within the source modality, we have
\begin{align}
\label{eqn:proof_upbound_sg_1}
\hE_{bs}(\{h, f\}) & \leq \hE_{s} (\{h, f\}) + \frac{1}{2} d_{\hH_s \Delta \hH_s} (\hD_s, \hD_{b}) + \lambda_s.
\end{align}
Then we can get
\begin{align}
\label{eqn:proof_upbound_sg_2}
\hE_{t} (\{\phi,g\})  \leq & \hE_{bt} (\{\phi,g\}) + \frac{1}{2}  d_{\hH_t \Delta \hH_t } (\hD_{b}, \hD_{t}) + \lambda_t \nonumber \\
= & \hE_{bs}(\{h, f\}) + (\hE_{bt} (\{\phi,g\}) - \hE_{bs}(\{h, f\}))\nonumber \\
&+ \frac{1}{2} d_{\hH_t \Delta \hH_t } (\hD_{b}, \hD_{t})  + \lambda_t \nonumber \\
\leq &\hE_{bs}(\{h, f\}) + \big| \hE_{bt} (\{\phi,g\}) - \hE_{bs}(\{h, f\}) \big| \nonumber \\
&+ \frac{1}{2}  d_{\hH_t \Delta \hH_t } (\hD_{b}, \hD_{t}) + \lambda_t,
\end{align}
where the first inequality holds due to Theorem 2 of \cite{ben2010theory} applied to the bridge and target domains within the target modality, and the second inequality is due to the property of the absolute value.  
According to the modality assumption, we have
\begin{align}
&\big| \hE_{bt} (\{\phi,g\}) - \hE_{bs}(\{h, f\}) \big|\nonumber \\
\leq & L \bE_{(\vx^{bs}, \vx^{bt}) \sim \hD_b} \big[d(h(\vx^{bs}),\phi(\vx^{bt})) \big].\label{eqn:proof_upbound_sg_3}
\end{align}
Based on the above three inequalities, we can have
\begin{align}
\label{eqn:proof_upbound_final}
\hE_{t} (\{\phi,g\})  \leq & \hE_{bs}(\{h, f\}) + \big| \hE_{bt} (\{\phi,g\}) - \hE_{bs}(\{h, f\}) \big| \nonumber \\
&+ \frac{1}{2} d_{\hH_t \Delta \hH_t } (\hD_{b}, \hD_{t})+ \lambda_t \nonumber \\
\leq &  \hE_{bs}(\{h, f\}) + L \bE_{(\vx^{bs}, \vx^{bt}) \sim \hD_b} \left[ d(h(\vx^{bs}),\phi(\vx^{bt})) \right] \nonumber \\
&+ \frac{1}{2}  d_{\hH_t \Delta \hH_t } (\hD_{b}, \hD_{t})  +  \lambda_t \nonumber \\
\leq & \hE_{s} (\{h, f\})  + L \bE_{(\vx^{bs}, \vx^{bt}) \sim \hD_b} \left[ d(h(\vx^{bs}), \phi(\vx^{bt})) \right] \nonumber \\        
& + \lambda_s + \lambda_t + \frac{1}{2} d_{\hH_s \Delta \hH_s} (\hD_s, \hD_{b}) \nonumber \\
&+ \frac{1}{2} d_{\hH_t \Delta \hH_t } (\hD_{b}, \hD_{t}),
\end{align}
where the three inequalities hold due to inequalities (\ref{eqn:proof_upbound_sg_2}), (\ref{eqn:proof_upbound_sg_3}), and (\ref{eqn:proof_upbound_sg_1}), respectively. Hence, we reach the conclusion.            
\end{proof}

Theorem \ref{the:unpperboundProof} shows that the target domain error is upper-bounded by the summation of five terms: 
(i) the source domain error $\hE_{s} (\{h, f\})$;
(ii) the modality discrepancy in the bridge domain;
(iii) the domain discrepancy between the source and bridge domains;
(iv) the domain discrepancy between the bridge and target domains; 
(v) the ideal combined errors, which are constants.
In LSB, term (i) is minimized by the segmentation loss defined in Eq.~(\ref{eqn:seg_loss_s}), term (ii) directly corresponds to the feature consistency loss $\hL_{\text{con}}^b$ defined in Eq.~(\ref{eqn:consistent_loss}), and the loss $\mathcal{L}_{\text{seg}}^{b}$ defined in Eq.~(\ref{eqn:seg_loss_b}) minimizes an empirical surrogate (with pseudo labels) of $\hE_{bt}$, the intermediate quantity through which the bound is derived. Terms (iii) and (iv) are fixed in the input space, but their counterparts on the learned representations are implicitly reduced by the alignment loss $\hL_{\text{ali}}$ defined in Eq.~(\ref{eqn:alignment_loss}), which, together with $\hL_{\text{con}}^b$, pulls the features of the three domains toward a shared structure~\cite{ganin2016domain}. Hence, the design of the LSB method aligns with the generalization bound in Theorem \ref{the:unpperboundProof}, which gives theoretical support for the proposed LSB method.

\section{Experiments}

In this section, we empirically evaluate the proposed LSB method under the HMUDA setting.

\label{app:ex_setting}
\begin{table}[!t]
    \centering
    \caption{Statistics of the number of samples in each split of datasets for all eight settings. Each setting is evaluated in both the 2D-to-3D and 3D-to-2D directions, which share the same data splits and differ only in which modality serves as the source domain. Note that the training samples in the target and bridge domains are without labels.}
    \resizebox{\linewidth}{!}{    
        \begin{NiceTabular}{l|cc|ccc}
        \toprule
        & \multicolumn{2}{c}{$\hB$} & $\hS$ & \multicolumn{2}{c}{$\hT$} \\
         \cmidrule(lr){2-3} \cmidrule(lr){4-4} \cmidrule(lr){5-6}
        Scenario & \multicolumn{2}{c}{Train} & Train & Train & Val/Test \\
        \midrule
        \textit{USA $\rightarrow$ Sing.}& \multirow{3}{*}{Sem.} &  \multirow{3}{*}{18,029}  & 15,695 & 9,665 & 2,770/2,929\\
        \textit{Day $\rightarrow$ Night}& & & 24,745 & 2,779 & 602/602\\
        \textit{Virt. $\rightarrow$ A2D2} & & & 2,126 & 24,461 & 808/2,426 \\
        \midrule
        \textit{USA $\rightarrow$ Sing.}& \multirow{3}{*}{A2D2} & \multirow{3}{*}{24,461}  & 15,695 & 9,665 & 2,770/2,929\\
        \textit{Day $\rightarrow$ Night}& & & 24,745 & 2,779 & 602/602\\
        \textit{Virt. $\rightarrow$ Sem.} & & & 2,126 & 18,029 & 1,101/4,071 \\
        \midrule
        \textit{Sem. $\rightarrow$ A2D2}& \multirow{2}{*}{Virt.} & \multirow{2}{*}{2,126}  & 18,029 & 24,461 & 808/2,426\\
        \textit{A2D2 $\rightarrow$ Sem.} & & & 24,461 & 18,029 & 1,101/4,071 \\
        \bottomrule
        \end{NiceTabular}
   }
    \label{tab:dataset_split}
\end{table}

\subsection{Experimental Settings}
\label{sec:set_up}
\noindent \textbf{Datasets.}
We conduct experiments on several publicly available multimodal datasets, including 
(i) nuScenes-lidarseg \cite{Caesar_2020_CVPR}, which is divided into different scene layouts (i.e., \textit{USA} and Singapore (\textit{Sing.})) and lighting conditions (i.e., \textit{Day} and \textit{Night}).
(ii) \textit{A2D2} \cite{geyer2020a2d2}, which consists of data collected from Audi, featuring diverse driving scenarios with multi-sensor data.
(iii) SemanticKITTI (\textit{Sem.}) \cite{Behley_2019_ICCV}, a large-scale dataset providing dense pointwise semantic annotations for LiDAR scans, capturing urban environments in various driving conditions.
(iv) VirtualKITTI (\textit{Virt.}) \cite{Gaidon_2016_CVPR}, a synthetic dataset generated from realistic 3D simulations with precise ground truth annotations.
To evaluate the proposed method under HMUDA, we construct six transfer tasks under the HMUDA setting, including:
(i) \textit{USA$\rightarrow$Sing.} with changes in scene layouts; (ii) \textit{Day$\rightarrow$Night} with changes in light conditions; (iii) \textit{Virt$\rightarrow$A2D2} and (iv) \textit{Virt$\rightarrow$Sem.} for synthetic to real data; 
(v) \textit{Sem.$\rightarrow$A2D2} and (vi) \textit{A2D2$\rightarrow$Sem.} with different camera setups.
For each task, the bridge domain is instantiated by a dataset disjoint from both the source and target domains. Accordingly, we use \textit{Sem.} for task (iii), \textit{A2D2} for task (iv), and \textit{Virt.} for tasks (v) and (vi). For tasks (i) and (ii), where both \textit{Sem.} and \textit{A2D2} satisfy this criterion, we evaluate each task with both choices of the bridge domain. This yields eight settings in total, as summarized in Table~\ref{tab:dataset_split}.
In all datasets, the LiDAR and RGB cameras are synchronized and calibrated.
To evaluate the generality of HMUDA, each of the eight settings is evaluated in both transfer directions, i.e., from 2D images to 3D point clouds 
(2D-to-3D) and from 3D point clouds to 2D images (3D-to-2D). 

\noindent \textbf{Implementation Details.}
By following \cite{xmuda_TPAMI},
for networks handling 2D image inputs, 
we use a U-Net~\cite{u-net} with a ResNet34~\cite{resnet} encoder pre-trained on ImageNet \cite{deng2009imagenet}.
For networks receiving 3D point clouds as the input, we use the SparseConvNet\cite{SCN} and U-Net\cite{u-net} architectures with a voxel size of 5cm. 
Moreover, we use a linear fully connected layer for projection networks $p_\phi$ and $p_h$, respectively.
Training is conducted using the Adam optimizer \cite{kingma2015adam} with a batch size of 16, an initial learning rate of 0.001, $\beta_1=0.9$, and $\beta_2=0.999$.
All the parameters are trained for 50,000 steps, with a learning rate scheduler following \cite{xmuda_cvpr}. 
The hyperparameter $\alpha$ in the EMA for constructing the teacher source model is initially set to 0.999. We set $\lambda_w=0.01$, $\lambda_c=4.0$, and $\lambda_a=0.1$ for all tasks.
We use the mean Intersection over Union (mIoU) to evaluate the performance.
All experiments are conducted on an NVIDIA V100 GPU with 32GB memory.

\noindent \textbf{Baselines.} 
We compare the proposed LSB method against the following representative baselines. 
\begin{enumerate*}[(i), series = tobecont, itemjoin = \quad]
\item \textbf{Oracle}, which trains the target network $(g,\phi)$ using segmentation loss directly on the labeled target domain $\{(\vx^t,\vy^t)\}$, where each input $\vx^t$ is paired with its ground-truth label $\vy^t$. This baseline serves as an upper bound for HMUDA.  

\item \textbf{xMUDA}~\cite{xmuda_TPAMI}, a multimodal UDA method that jointly learns from paired 2D-3D data in both the source and target domains by enforcing cross-modal consistency through mutual mimicking.
Such paired supervision provides stronger guidance than unimodal settings but requires costly annotations. 
We therefore regard xMUDA as another \emph{upper-bound} baseline for HMUDA.  

\item \textbf{Source-Only}, where the source network is trained solely on the source domain $\hS$, and pseudo-labels generated on the bridge domain $\hB$ are then used to train the target network.  

\item \textbf{Pseudo-Labeling (PL)}~\cite{PL}, a two-stage baseline implemented following~\cite{xmuda_TPAMI}. The source network trained on the source domain generates pseudo-labels for source-modality samples in the bridge domain. These labels are used as the supervision information to train the target network on the target-modality samples in the bridge domain. After that, the trained target network is used to make prediction on data in the target domain during the testing process. 

\item \textbf{CDSPP}~\cite{wang2022cross}, an HDA method that transfers knowledge from extracted features by learning domain-specific projections to align source and target domains in a shared subspace while preserving the class structure, which we reproduce using approximately 5\% labeled target samples in a semi-supervised setup.
For a fair comparison, we use the same backbone ResNet34 and SparseConvNet for 2D and 3D samples as LSB to extract features. 

\end{enumerate*}
Note that CDSPP is allowed to access a small set of labeled target samples, whereas all other baselines, including the proposed LSB, do not use any labeled target data and strictly follow the unsupervised setting for the target domain.

\begin{table*}[!t]
\centering
\caption{Testing results on HMUDA tasks in terms of mIoU. Each domain transfer is evaluated in both 2D-to-3D and 3D-to-2D modality directions. The last column reports the average over all eight settings for each transfer direction. The best performance is in \textbf{bold}.}
    \setlength{\tabcolsep}{3pt}
    \resizebox{1\textwidth}{!}{
        \begin{NiceTabular}{l|l|ccc|ccc|cc|c}
        \toprule

        &  & \textit{USA $\rightarrow$ Sing.} & \textit{Day $\rightarrow$ Night} & \textit{Virt. $\rightarrow$ A2D2} & \textit{USA $\rightarrow$ Sing.} & \textit{Day $\rightarrow$ Night} & \textit{Virt. $\rightarrow$ Sem.} & \textit{Sem. $\rightarrow$ A2D2} & \textit{A2D2 $\rightarrow$ Sem.} & Avg. \\

       \midrule

         & Bridge Domain {$\hB$} & \multicolumn{3}{c}{\textit{Sem.}} &\multicolumn{3}{c}{\textit{A2D2}} & \multicolumn{2}{c}{\textit{Virt.}} & -- \\
        \midrule
        \midrule
        \multirow{6}{*}{\rotatebox{90}{2D\textit{$\rightarrow$}3D}} & Oracle & 77.69 & 73.71 & 71.50 & 77.69 & 73.71 & 82.75 & 71.50 & 82.75 & 76.41 \\ 
        & xMUDA~\cite{xmuda_TPAMI} & 65.12 & 75.11 & 61.03 & 65.12 & 75.11 & 57.97 & 44.95 & 65.07 & 63.69\\
        \cmidrule(){2-11}
        & Source-Only & 51.04 & 57.32 & 19.00 & 49.21 & 55.96 & 36.44 & 43.75 & 43.28 & 44.50 \\
        & PL~\cite{PL} & 52.08 & 59.33 & 16.84 & 55.39 & 61.95 & 44.49 & 39.66 & 38.25 & 46.00\\
        & CDSPP~\cite{wang2022cross} & 17.61 & 21.81 & 9.74 & 17.61 & 21.81 & 12.44 & 11.27 & 11.00 & 15.41 \\
        \rowcolor{Gray}
        & LSB & \textbf{56.41} & \textbf{62.25} & \textbf{33.37} & \textbf{57.13} & \textbf{63.15} & \textbf{45.43} & \textbf{46.34} & \textbf{47.22} & \textbf{51.41}\\
        
    \midrule \midrule

        \multirow{6}{*}{\rotatebox{90}{3D\textit{$\rightarrow$}2D}} & Oracle & 76.28 & 62.63 & 85.67 & 76.28 & 62.63 & 87.19 & 85.67 & 87.19 & 77.94 \\   
        & xMUDA~\cite{xmuda_TPAMI} & 65.88 & 63.63 & 62.07 & 65.88 & 63.63 & 52.98 & 63.54 & 60.51 & 62.27\\
        \cmidrule(){2-11}
        & Source-Only & 43.80 & 17.69 & 42.56 & 38.99 & 25.02 & 39.92 & 31.48 & 20.06 & 32.44 \\
        & PL~\cite{PL} & 43.89 & 26.55 & 43.37 & 34.97 & 22.67 & 38.75 & 14.45 & 34.27 & 32.37 \\
        & CDSPP~\cite{wang2022cross} & 22.18 & 17.75 & 20.53 & 22.18 & 17.75 & 15.87 & 17.58 & 13.11 & 18.37 \\        
        \rowcolor{Gray}
        & LSB & \textbf{45.35} & \textbf{33.81} & \textbf{47.92} & \textbf{39.83} & \textbf{38.36} & \textbf{40.03} & \textbf{32.49} & \textbf{39.89} & \textbf{39.71}\\     
        \bottomrule
        \end{NiceTabular}
   }
    \label{tab:main_result_label}
\end{table*}

\subsection{Main Results}

Table \ref{tab:main_result_label} reports the mIoU results on eight transfer settings in both the 2D-to-3D and 3D-to-2D directions. LSB consistently outperforms Source-Only, PL, and CDSPP across all settings in both directions, demonstrating its effectiveness in transferring semantic knowledge across domains and modalities through an unlabeled bridge domain.
For example, on the 2D-to-3D task \textit{Virt.$\rightarrow$A2D2} with bridge domain \textit{Sem.}, LSB surpasses Source-Only, PL, and CDSPP by a large margin of $14.37$, $16.53$, and $23.63$, respectively, validating its effectiveness in addressing the modality gap across domains. 
LSB achieves average mIoU improvements of $5.41$ and $7.35$ over PL across all the 2D-to-3D and 3D-to-2D tasks, respectively, demonstrating its capability to bridge source and target networks through joint training.
Moreover, LSB consistently outperforms CDSPP across all tasks, achieving average mIoU gains of 36.00 and 21.34 on the 2D-to-3D and 3D-to-2D settings, respectively, demonstrating the effectiveness of explicitly exploiting instance-level cross-modal correspondence within an end-to-end training framework.
It is worth noting that CDSPP utilizes about 5\% labeled target samples, whereas LSB does not rely on any labeled target data, indicating that exploiting an unlabeled bridge domain can be competitive with few-shot HDA methods.
Compared with the supervision-heavy xMUDA, which leverages paired 2D–3D data in both source and target domains,
LSB achieves a higher mIoU on the 2D-to-3D transfer task \textit{Sem.$\rightarrow$A2D2} with the bridge domain \textit{Virt.}, suggesting that the bridge domain can be effective for knowledge transfer even with less direct supervision. 
Despite being somewhat effective, all methods still fall short of the Oracle, underscoring the challenges inherent in HMUDA tasks.

\begin{table*}
    \centering
    \caption{Effect of losses $\hL_{seg}^s$, $\hL_{seg}^b$, $\hL_{con}^b$, and $\hL_{ali}$ in terms of mIoU for 2D-to-3D HMUDA tasks. The bridge domain $\hB$ for each task is given in parentheses. The best performance is in \textbf{bold}.}
    \label{tab:ablation_label}
    \resizebox{1\textwidth}{!}{    
        \begin{NiceTabular}{cccc|cccc}
        \toprule
        $\hL_{seg}^s$ & $\hL_{seg}^b$ & $\hL_{con}^b$ & $\hL_{ali}$ & \textit{USA $\rightarrow$ Sing.} (\textit{Sem.}) & \textit{Day $\rightarrow$ Night} (\textit{Sem.}) & \textit{Virt. $\rightarrow$ A2D2} (\textit{Sem.}) & \textit{Sem. $\rightarrow$ A2D2} (\textit{Virt.}) \\
        \midrule
        \cmark & \cmark & \xmark & \xmark & 53.62 & 61.86 & 22.72 & 42.47 \\
        \cmark & \cmark & \cmark & \xmark & 52.68 & 61.12 & 28.24 & 45.27 \\        
        \cmark & \cmark & \xmark & \cmark & 54.34 & 60.86 & 27.96 & \textbf{46.37} \\
        \rowcolor{Gray}
        \cmark & \cmark & \cmark & \cmark &  \textbf{56.41} & \textbf{62.25} & \textbf{33.37} & 46.34 \\ 
        \midrule \midrule
        $\hL_{seg}^s$ & $\hL_{seg}^b$ & $\hL_{con}^b$ & $\hL_{ali}$ & \textit{USA $\rightarrow$ Sing.} (\textit{A2D2}) & \textit{Day $\rightarrow$ Night} (\textit{A2D2}) & \textit{Virt. $\rightarrow$ Sem.} (\textit{A2D2}) & \textit{A2D2 $\rightarrow$  Sem.} (\textit{Virt.}) \\
        \midrule 
        \cmark & \cmark & \xmark & \xmark & 49.35 & 60.64 & 35.74 & 42.29 \\
        \cmark & \cmark & \cmark & \xmark & 50.50 & 62.72 & 36.53 & 42.46 \\        
        \cmark & \cmark & \xmark & \cmark & 54.49 & 62.87 & 38.55 & 46.95 \\
        \rowcolor{Gray}
        \cmark & \cmark & \cmark & \cmark & \textbf{57.13} & \textbf{63.15} & \textbf{45.43} & \textbf{47.22} \\ 
        \bottomrule
        \end{NiceTabular}
    }
\end{table*}

\subsection{Ablation Studies}

\noindent {\bf Effect of Different Losses
.}
We conduct experiments on HMUDA tasks to study the effect of losses, i.e., $\hL_{seg}^s$, $\hL_{seg}^b$, $\hL_{con}^b$, and $\hL_{ali}$, in terms of mIoU.
Specifically, we consider four combinations: 
(i) with the two segmentation losses $\hL_{seg}^s$ and $\hL_{seg}^b$ only (i.e., LSB (w/ seg));
(ii) without the alignment loss $\hL_{ali}$ (i.e., LSB (w/o $\hL_{ali}$)); 
(iii) without the feature consistency loss $\hL_{con}^b$ (i.e., LSB (w/o $\hL_{con}^b$)); 
(iv) with all the proposed losses $\hL_{seg}^s$, $\hL_{seg}^b$, $\hL_{con}^b$ and $\hL_{ali}$ together (i.e. the proposed LSB method).
Since the two segmentation losses (i.e., $\hL_{seg}^s$ and $\hL_{seg}^b$) are necessary to train the source and target networks, we do not remove one of them in the aforementioned variants of LSB.

Table \ref{tab:ablation_label} shows the testing results on the 2D-to-3D HMUDA tasks for different variants of LSB. 
As can be seen, LSB achieves the best mIoU across all HMUDA tasks, except for \textit{Sem.$\rightarrow$A2D2} where it performs comparable with LSB (w/o $\hL_{con}^b$). 
LSB outperforms LSB (w/o $\hL_{con}^b$) by a large margin of $2.36$ on average, validating that encouraging the consistent features benefits the learning of the target domain.
Compared with LSB (w/o $\hL_{ali}$), LSB achieves an average improvement of $3.97$, showing the effectiveness of minimizing the distance of class centroids between the source and target domains.
Moreover, LSB (w/ seg) surpasses the state-of-the-art UDA method (i.e., PL in Table \ref{tab:main_result_label}) by $1.19$ on average, demonstrating the superiority of the joint training strategy. 
In conclusion, this ablation study demonstrates the usefulness and effectiveness of the four losses proposed in LSB.

\noindent \textbf{Effect of $p_h$ and $p_\phi$.}
In LSB, we introduce two projection networks to map the source and target features into a $d$-dimensional shared feature space.
To study the effect of these two projection networks, we compare LSB with some variants: 
(i) LSB (w/o $p_h$), which uses a linear fully-connected layer to map the target feature into the $d_h$-dimensional source feature space, where $d_h$ is the output dimension of the source encoder $h$ (Section~\ref{sec:architecture}); 
(ii) LSB (w/o $p_\phi$), which uses a linear fully-connected layer to map the source feature into the $d_\phi$-dimensional target feature space, where $d_\phi$ is the output dimension of the target feature extractor $\phi$ (Section~\ref{sec:architecture}). 
Usually $d_h$ is different from $d_\phi$, and hence we cannot build a variant without both $p_h$ and $p_\phi$.

Table \ref{tab:abl_projection} shows the testing results for three 2D-to-3D HMUDA tasks. 
As can be seen, LSB consistently outperforms LSB (w/o $p_\phi$) and LSB (w/o $p_h$) across all the tasks, validating that aligning the source and target domains in a common latent space helps the effective transfer.
Notably, LSB (w/o $p_\phi$) performs significantly worse, indicating that directly aligning source samples to the target feature space hinders the learning of segmentation.

\begin{figure*}[!t]
    \centering
    \subfloat[Effect of $\lambda_c$.]{%
        \includegraphics[height=3.0cm, keepaspectratio]{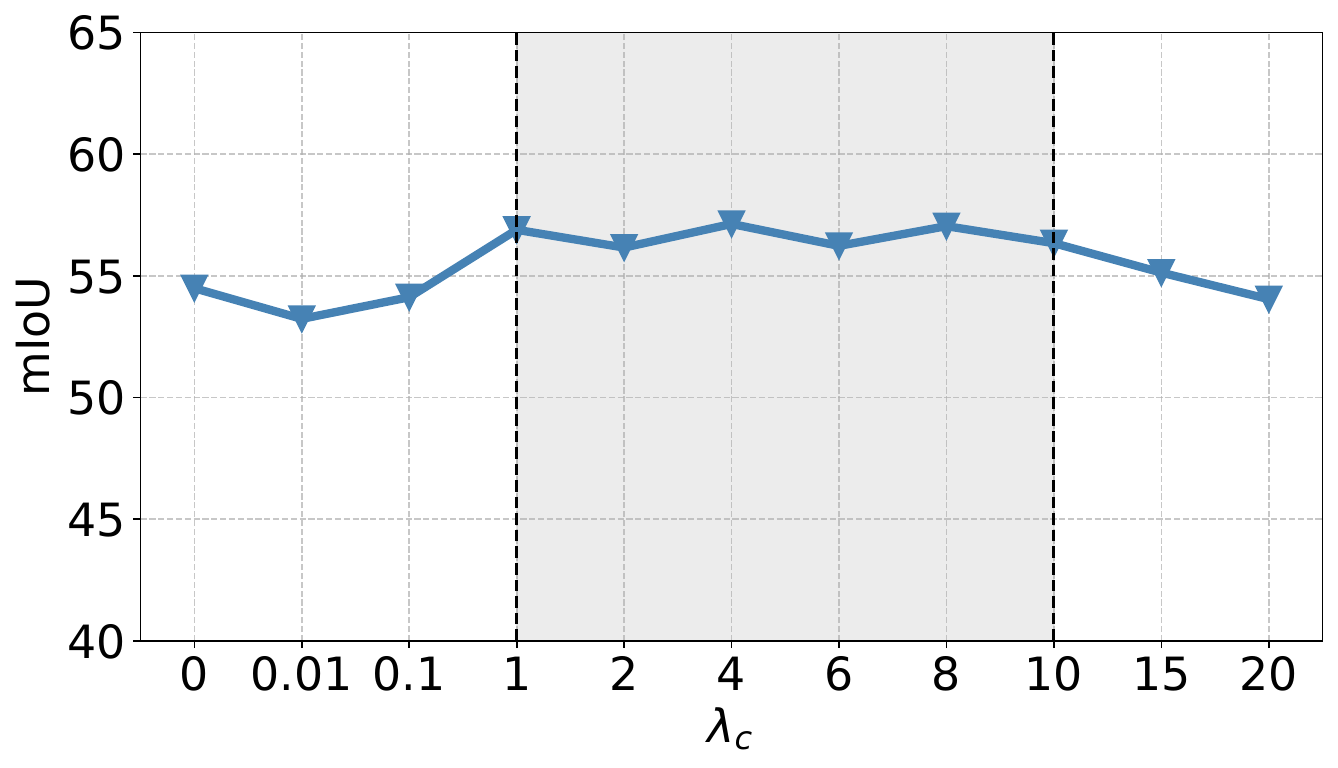}%
        \label{fig:sen_lambda_c}%
    }
    \hfill
    \subfloat[Effect of $\lambda_a$.]{%
        \includegraphics[height=3.0cm, keepaspectratio]{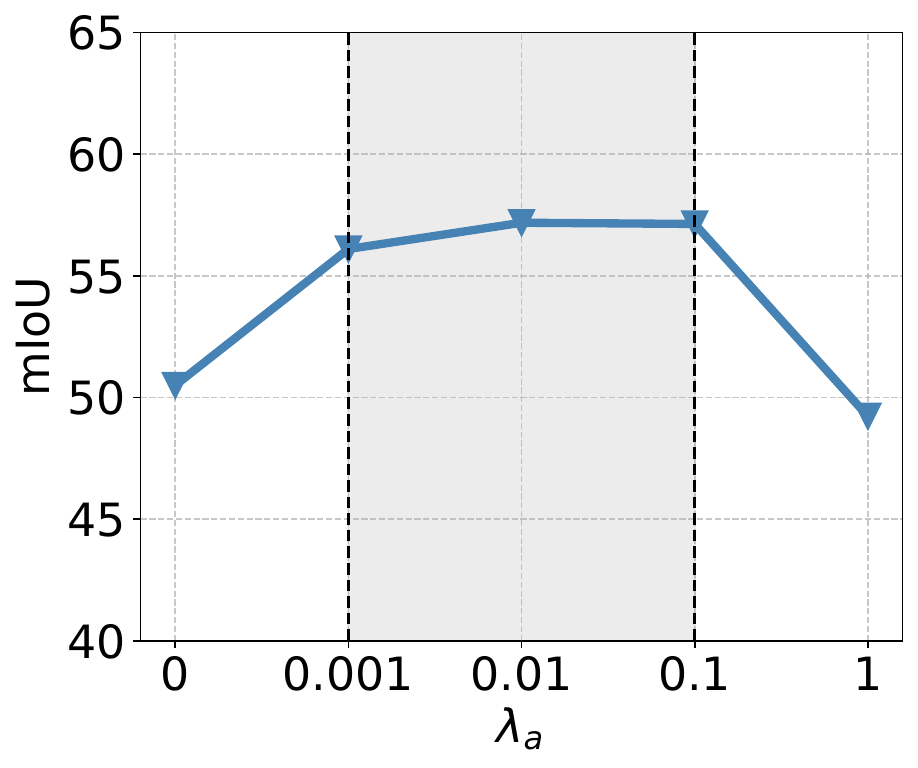}%
        \label{fig:sen_lambda_a}%
    }
    \hfill
    \subfloat[Effect of $\lambda_w$.]{%
        \includegraphics[height=3.0cm, keepaspectratio]{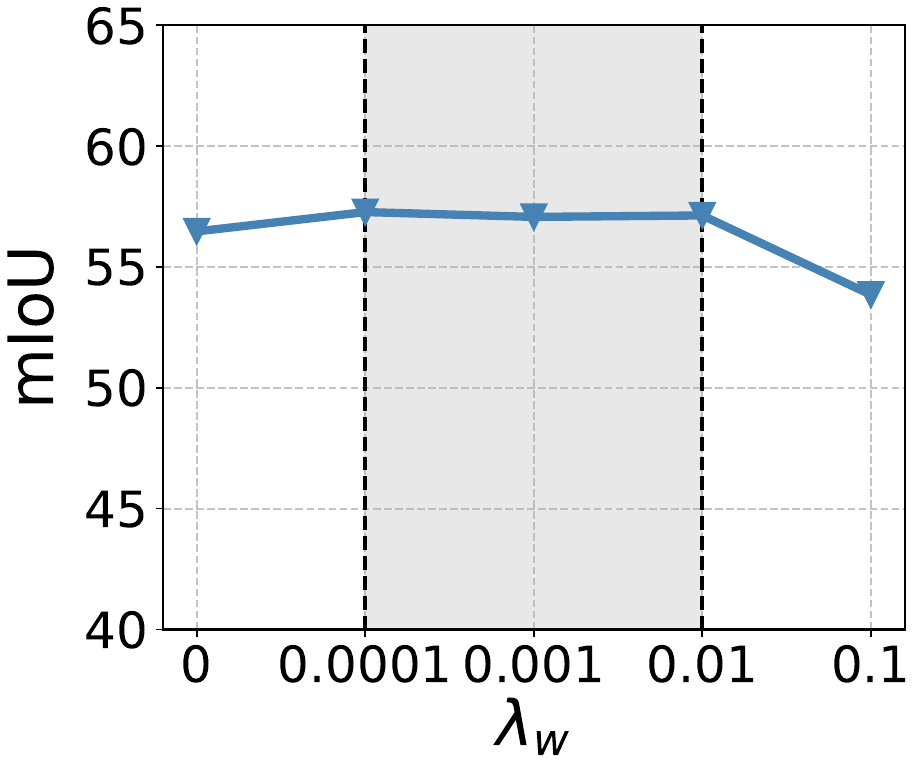}%
        \label{fig:sen_lambda_w}%
    }
    \hfill
    \subfloat[Effect of the number of bridge samples.]{%
        \includegraphics[height=3.0cm, keepaspectratio]{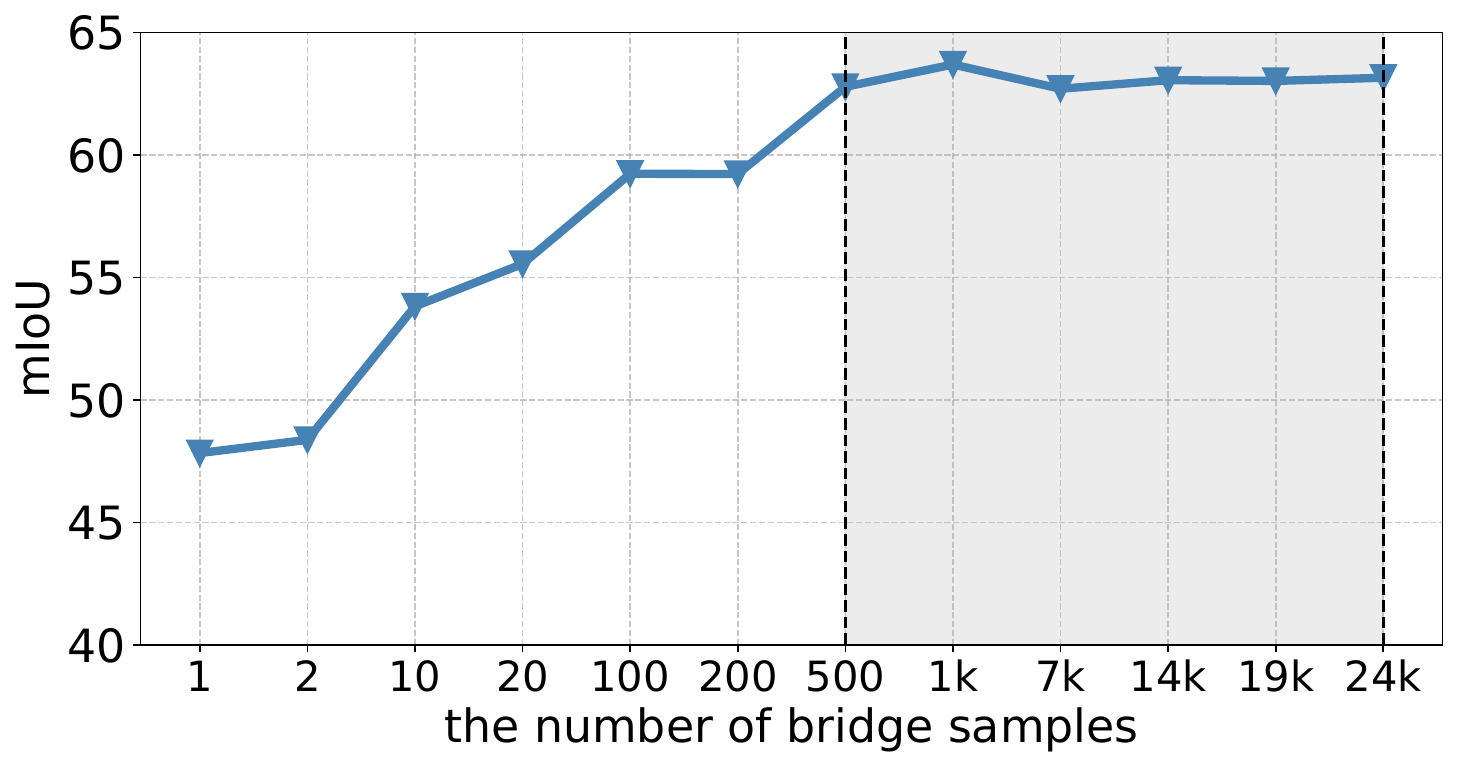}%
        \label{fig:sen_sample_size}%
    }
    \caption{Sensitivity analysis of the LSB method with respect to hyperparameters and the number of samples in the bridge domain.}
    \label{fig:sensitivity}
\end{figure*}

\noindent \textbf{Effect of cross-modal alignment.}
We study the effect of cross-modal domain alignment between the source and target domain on the HMUDA tasks.
Instead of aligning the target domain with the source domain, 
we conduct an additional experiment, i.e., LSB (\textit{w.} $\hL_{ali}(\hB, \hT)$), by minimizing the class centroids between the bridge and target domains. 
These class centroids are computed using pseudo labels for the bridge and target domains in a way similar to Eqs. (\ref{centroid_source_domain}) and (\ref{centroid_target_domain}).
As shown in Table \ref{tab:alignment}, LSB consistently outperforms LSB (\textit{w.} $\hL_{ali}(\hB, \hT)$), indicating that aligning the target domain with the source domain is more effective.

\begin{table}
  \centering
  \caption{Effect of $p_\phi$ and $p_h$ on three 2D-to-3D HMUDA tasks.}
  \resizebox{1\linewidth}{!}{
        \begin{NiceTabular}{l|c|cc|c|c}
        \toprule
         &  & \textit{USA} & \textit{Day} & & \textit{A2D2} \\

        Method & $\hB$ & $\rightarrow$ & $\rightarrow$ & $\hB$ & $\rightarrow$ \\

        &  & \textit{Sing.} & \textit{Night} & & \textit{Sem.} \\
        \midrule
        LSB (w/o $p_\phi$) & {\multirow{3}{*}{\textit{A2D2}}} & 20.78 & 3.82 & {\multirow{3}{*}{\textit{Virt.}}} & 15.00 \\ 
        LSB (w/o $p_h$) & & 40.47 & 43.10 & & 32.52 \\ 
        \rowcolor{Gray}
        LSB & & 57.13 & 63.15 & & 47.22 \\       
        \bottomrule
        \end{NiceTabular}
   }
    \label{tab:abl_projection}
\end{table}

\begin{table}
    \centering
    \caption{Effect of cross-modal alignment on three 2D-to-3D HMUDA tasks.}    
    \resizebox{1\linewidth}{!}{    
        \begin{NiceTabular}{l|c|cc|c|c}
        \toprule
         &  & \textit{USA} & \textit{Day} &  & \textit{A2D2} \\
        Method & $\hB$ & $\rightarrow$ & $\rightarrow$ & $\hB$ & $\rightarrow$ \\
         &  & \textit{Sing.} & \textit{Night} &  & \textit{Sem.} \\
        \midrule
        LSB (\textit{w.} $\hL_{ali}(\hB, \hT)$) & {\multirow{2}{*}{\textit{A2D2}}} & 50.86 & 60.12 & {\multirow{2}{*}{\textit{Virt.}}} & 44.44 \\ 
        LSB & & 57.13 & 63.15 & & 47.22 \\       
        \bottomrule
        \end{NiceTabular}
   }
    \label{tab:alignment}
\end{table}

\subsection{Sensitivity Analysis}

\begin{figure*}[!t]
    \centering
    \includegraphics[width=1\linewidth]{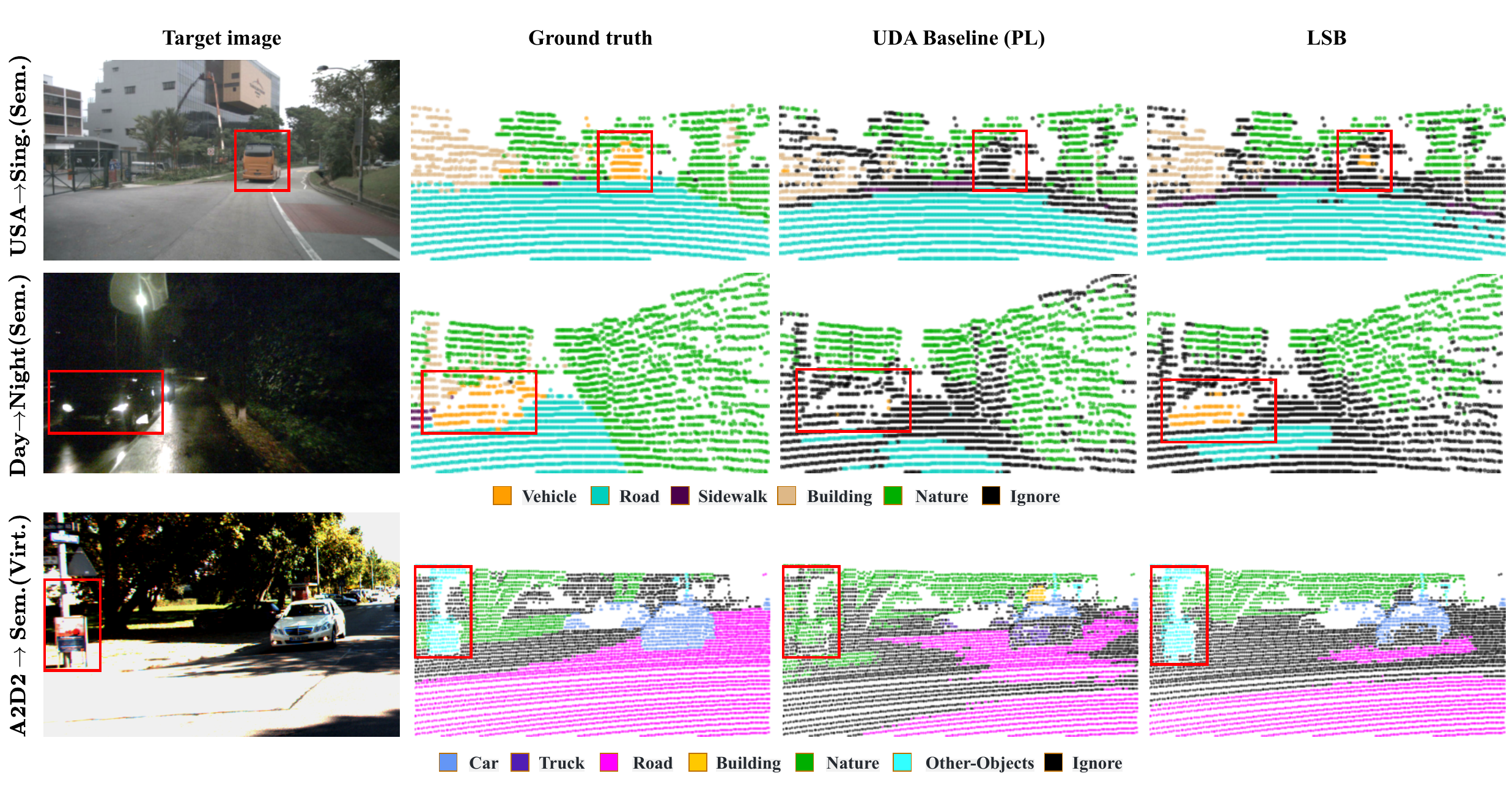}
    \caption{Qualitative results on three 2D-to-3D HMUDA tasks: \textit{USA$\rightarrow$Sing.}, \textit{Day$\rightarrow$Night}, and \textit{A2D2$\rightarrow$Sem.}. $(\cdot)$ in the vertical axis denotes the bridge domain $\hB$ used in the HMUDA task. For example, \textbf{USA$\rightarrow$Sing.(Sem.)} denotes the transfer from \textit{USA} to \textit{Sing.} via the bridge domain \textit{Sem.}.}
    \label{fig:visual}
\end{figure*}

In this section, we evaluate the sensitivity of the performance with respect to hyperparameters and the number of bridge samples.

\noindent \textbf{Effect of $\lambda_c$.} 
We conduct experiments on a 2D-to-3D transfer task of \textit{USA$\rightarrow$Sing.} via the bridge domain \textit{A2D2} to study the effect of $\lambda_c$ to the performance. 
As shown in Figure~\ref{fig:sen_lambda_c}, setting $\lambda_c$ to $0.01$ or $0.1$ yields no improvement over the baseline corresponding to $\lambda_c=0$. Increasing $\lambda_c$ to $1$ improves the performance, and the performance remains stable throughout a range $\lambda_c \in [1,10]$. Larger weights than 10 gradually degrade the performance, indicating that the feature consistency loss is beneficial at a moderate weight but may dominate the entire objective when over-weighted.

\noindent {\bf Effect of $\lambda_a$.}
To investigate the sensitivity of the performance with respect to $\lambda_a$, we conduct experiments on a 2D-to-3D transfer task of \textit{USA$\rightarrow$Sing.} with the bridge domain \textit{A2D2}. 
As shown in Figure~\ref{fig:sen_lambda_a}, setting $\lambda_a=0.001$ substantially improves mIoU over $\lambda_a=0$. The performance remains stable for $\lambda_a \in [0.01,0.1]$, while increasing it to $1$ causes a marked drop. These results show that a moderate weight for the alignment loss is effective, whereas an excessively large value can over-constrain the learned representations and impair the performance.

\noindent \textbf{Effect of $\lambda_w$.}
To investigate the sensitivity of the performance with respect to $\lambda_w$, we conduct experiments on the 2D-to-3D transfer task \textit{USA$\rightarrow$Sing.} with the bridge domain \textit{A2D2}. As shown in Figure~\ref{fig:sen_lambda_w}, introducing a small regularization weight of $10^{-4}$ improves the mIoU over $\lambda_w=0$. The performance remains stable for $\lambda_w\in[10^{-4},10^{-2}]$, while increasing the weight to $10^{-1}$ causes a performance drop. These results show that a moderate regularization stabilizes the projection networks, whereas an excessively large value over-constrains the learned projections.

\begin{figure*}
    \centering
    \subfloat[Source-Only]{
        \includegraphics[width=0.45\linewidth]{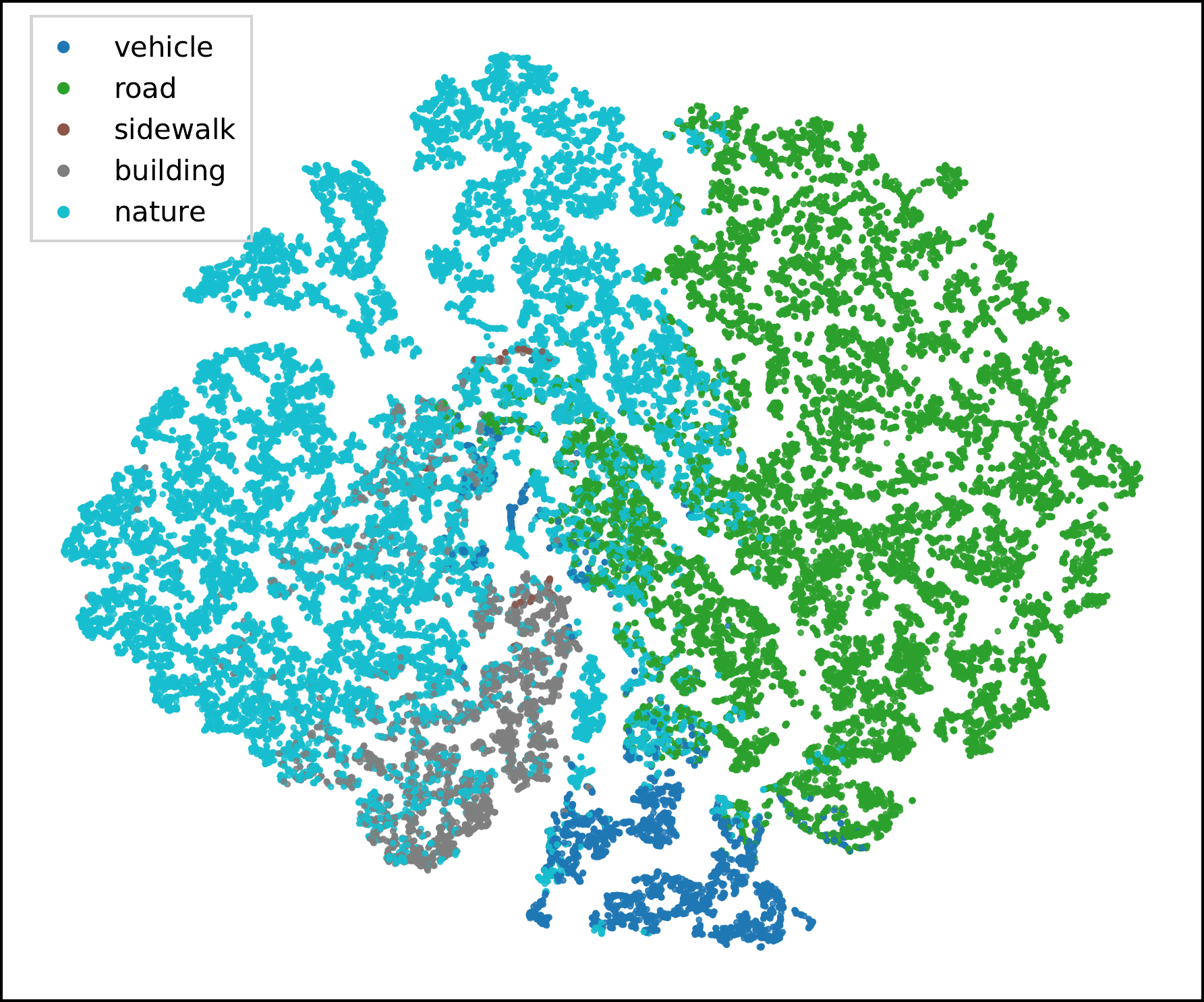}
        \label{fig:tsne_source_only}
    }
    \hfill
    \subfloat[LSB]{
        \includegraphics[width=0.45\linewidth]{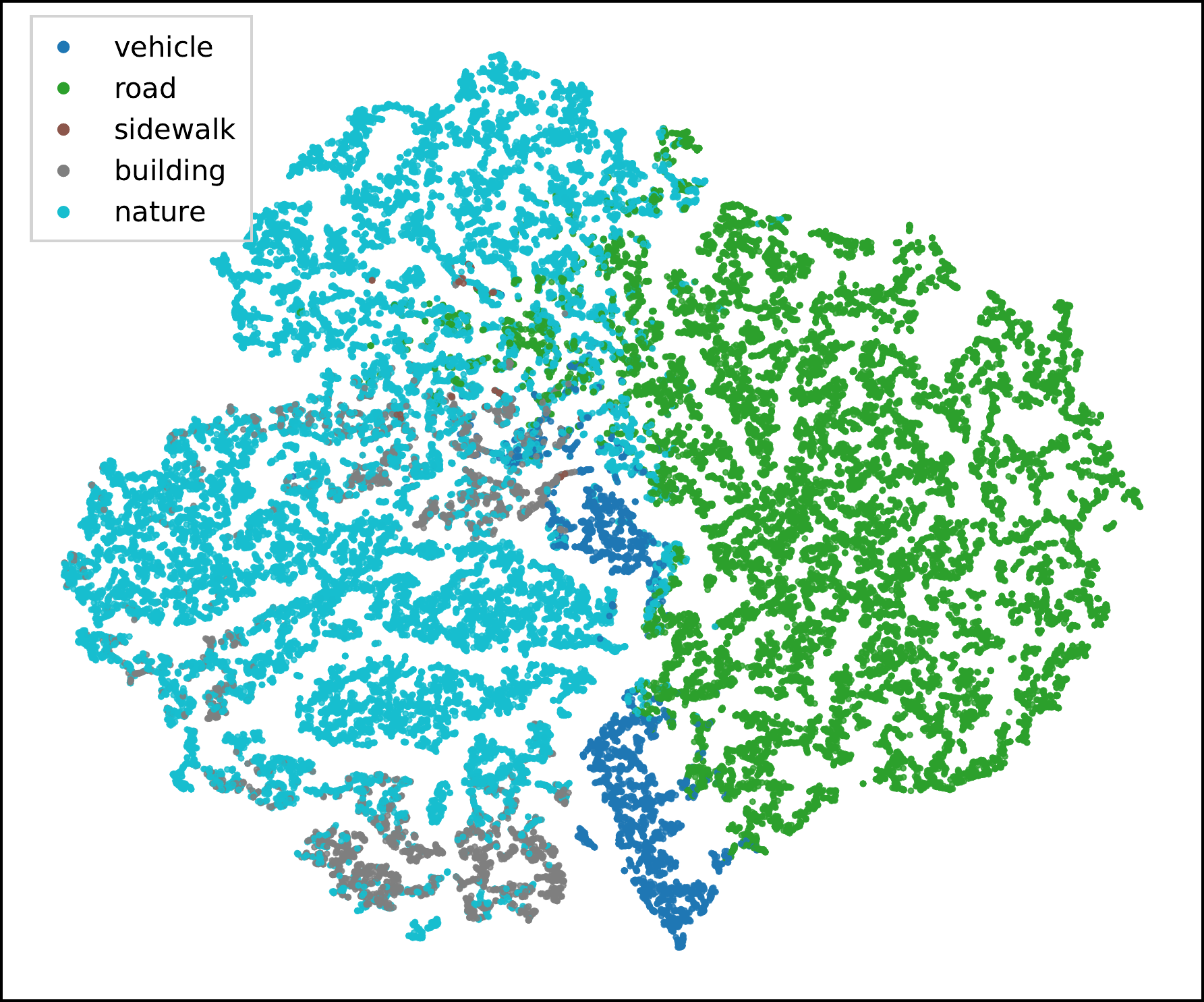}
        \label{fig:tsne_LSB}
    }
    \caption{t-SNE visualization of target feature embeddings for the Source-Only and LSB methods.}
    \label{fig:tsne}
\end{figure*}

\noindent {\bf Effect of the number of bridge samples.}
To study the sensitivity of the performance with respect to the number of samples in the bridge domain, We conduct experiments on a  2D-to-3D transfer task of \textit{Day$\rightarrow$Night} with the bridge domain \textit{A2D2}. As shown in Figure \ref{fig:sen_sample_size}, mIoU increases with the number of bridge samples and stabilizes after 500 samples, suggesting that a relatively small number of bridge samples is sufficient for LSB to achieve stable and good performance.

\subsection{Visualization}
\label{app:Visualization}

\noindent {\bf Qualitative results.}
Figure~\ref{fig:visual} presents the qualitative segmentation results for various 2D-to-3D tasks, including \textit{USA$\rightarrow$Sing.}, \textit{Day$\rightarrow$Night}, and \textit{A2D2$\rightarrow$Sem}.
As can be seen, the LSB method predicts segmented objects more accurately than the PL method.
For instance, in the image from the task of USA$\rightarrow$Sing., we can see that the proposed LSB method correctly identifies the bus that is highlighted with a red bounding box, while the PL method misclassifies it as `Sidewalk'. 
Similar improvements are observed in the images from tasks of \textit{Day$\rightarrow$Night} and \textit{A2D2$\rightarrow$Sem}, where LSB correctly recognizes them as `Vehicle' and `Other-Objects'.

\noindent {\bf t-SNE Visualization.}
Figure~\ref{fig:tsne} presents the t-SNE visualization of target feature embeddings from the Source-Only and LSB methods on the 2D-to-3D task of \textit{Day$\rightarrow$Night} via the bridge domain \textit{A2D2}. Compared with the Source-Only method, the LSB method shows improved clustering results for the `vehicle' class, with more compact and better separated features for the target domain.

\section{Conclusion}

In this paper, we introduce HMUDA, a novel setting designed to transfer knowledge between heterogeneous modalities using a bridge domain. 
For HMUDA, we propose a specialized framework LSB, with two distinct networks tailored for the source and target modalities.
These networks are trained jointly with two segmentation losses specific to each network, alongside the feature consistency loss to promote similar feature representations between domains and the domain alignment loss to reduce domain discrepancies. 
Experimental results on various HMUDA benchmark datasets demonstrate the effectiveness of LSB in transferring knowledge across heterogeneous modalities. 
Although our experiments focus on 2D images and 3D point clouds, the HMUDA formulation and LSB architecture are general and can in principle be applied to other modality pairs, such as RGB image--infrared image, audio--text, or multi-spectral image--RGB image, which could be one of our future work.

\section*{Acknowledgements}

This work was supported by National Key Research and Development Program of China under Grant no. 2022YFA1004102, National Natural Science Foundation of China under Grant no. 62136005, and Shenzhen fundamental research program JCYJ20250604144724032.

{
    \small
    \bibliographystyle{IEEEtran}
    \bibliography{ref}
}

\clearpage

\vfill

\end{document}

%% file: math_commands.tex
\usepackage[pagebackref=false,breaklinks=true,colorlinks,bookmarks=false]{hyperref}
\hypersetup{linkcolor=[rgb]{0.7,0.1,0.1}}
\hypersetup{citecolor=[rgb]{0.4,0.15,0.95}}

\usepackage[capitalize,noabbrev]{cleveref}
\usepackage{colortbl}
\usepackage{nicematrix}
\usepackage{pifont}
\newcommand{\cmark}{\ding{51}}%
\newcommand{\xmark}{\ding{55}}%
\usepackage{xcolor,colortbl}

\theoremstyle{plain}
\newtheorem{theorem}{Theorem}

\newtheorem{definition}{Definition}

\theoremstyle{remark}

\usepackage{algorithm}

\definecolor{Gray}{gray}{0.94}

\DeclareMathOperator*{\argmin}{arg\,min}

\newcommand{\bE}{\mathbb{E}}

\newcommand{\hB}{\mathcal{B}}

\newcommand{\hD}{\mathcal{D}}
\newcommand{\hE}{\mathcal{E}}

\newcommand{\hH}{\mathcal{H}}

\newcommand{\hL}{\mathcal{L}}
\newcommand{\hM}{\mathcal{M}}

\newcommand{\hS}{\mathcal{S}}
\newcommand{\hT}{\mathcal{T}}

\newcommand{\vm}{{\bf m}}

\newcommand{\vq}{{\bf q}}

\newcommand{\vw}{{\bf w}}
\newcommand{\vx}{{\bf x}}
\newcommand{\vy}{{\bf y}}

\newcommand{\vtheta}{{\boldsymbol \theta}}

\usepackage[inline, shortlabels]{enumitem}